\documentclass[10pt,twocolumn,letterpaper]{article}

\usepackage[pagenumbers]{cvpr} 

\definecolor{cvprblue}{rgb}{0.21,0.49,0.74}
\usepackage[pagebackref,breaklinks,colorlinks,allcolors=cvprblue]{hyperref}

\title{SL-YOLO: A Stronger and Lighter Drone Target Detection Model}

\author{Defan Chen\\
Shenzhen University\\
\and
Luchan Zhang\\
Shenzhen University\\
}

\begin{document}
\maketitle
\begin{abstract}
Detecting small objects in complex scenes, such as those captured by drones, is a daunting challenge due to the difficulty in capturing the complex features of small targets. While the YOLO family has achieved great success in large target detection, its performance is less than satisfactory when faced with small targets. Because of this, this paper proposes a revolutionary model SL-YOLO (Stronger and Lighter YOLO) that aims to break the bottleneck of small target detection. We propose the Hierarchical Extended Path Aggregation Network (HEPAN), a pioneering cross-scale feature fusion method that can ensure unparalleled detection accuracy even in the most challenging environments. At the same time, without sacrificing detection capabilities, we design the C2fDCB lightweight module and add the SCDown downsampling module to greatly reduce the model's parameters and computational complexity. Our experimental results on the VisDrone2019 dataset reveal a significant improvement in performance, with mAP$_{0.5}$ jumping from 43.0\% to 46.9\% and mAP$_{0.5:0.95}$ increasing from 26.0\% to 28.9\%. At the same time, the model parameters are reduced from 11.1M to 9.6M, and the FPS can reach 132, making it an ideal solution for real-time small object detection in resource-constrained environments.
\end{abstract}

\section{Introduction}
\label{sec:intro}

As drone technology rapidly advances, aerial photography has become an essential tool in critical fields such as disaster monitoring, traffic management, search and rescue, and agricultural oversight. Unlike traditional ground-based methods, drone imagery offers a high-altitude perspective, wide coverage, and reduced operational costs. However, small-target detection in drone imagery presents significant challenges, including complex backgrounds, low-contrast objects, and dynamic environmental conditions, often leading to detection errors and missed targets~\cite{wu2021deep,tang2023survey}. Achieving real-time, accurate detection is essential for enabling drones to perform effectively across various applications. For instance, in disaster response, precise detection can identify survivors or hazards, while in agriculture, detecting small anomalies can lead to better crop management. By addressing these limitations, drones can operate reliably and efficiently, maximizing their potential in complex and demanding conditions.

Existing target detection methods are broadly categorized into two approaches: the R-CNN series two-stage models~\cite{girshick2014rich,girshickICCV15fastrcnn,renNIPS15fasterrcnn}, and the single-stage YOLO series~\cite{redmon2016you,redmon2017yolo9000,farhadi2018yolov3,bochkovskiy2020yolov4,li2022yolov6,wang2023yolov7,wang2024yolov9,wang2024yolov10,Jocher_Ultralytics_YOLO_2023}. YOLOv8 achieves a strong balance between speed and accuracy, making it popular in many applications. Yet, it still struggles with small target detection in complex scenes.
Many recent studies have attempted to tackle this, employing multi-scale feature fusion, attention mechanisms~\cite{vaswani2017attention}, and lightweight designs. Classic approaches like Feature Pyramid Networks (FPN)\cite{lin2017feature}, Path Aggregation Networks (PAN)\cite{liu2018path}, and the more recent BiFPN~\cite{tan2020efficientdet} optimize multi-scale feature processing. However, even with these advancements, small-target detection in drone images remains challenging. Lightweight models such as MobileNets~\cite{howard2017mobilenets,sandler2018mobilenetv2,howard2019searching}, ShuffleNets \cite{zhang2018shufflenet,ma2018shufflenet}, and EfficientNets\cite{tan2019efficientnet,tan2021efficientnetv2} reduce computational cost while maintaining accuracy, but they fall short in drone-based small-target scenarios.

By overcoming the limitations of existing models, our study proposes a stronger yet lightweight model, SL-YOLO, which builds on YOLOv8s to handle small target detection under complex conditions and on resource-constrained devices. This model not only breaks the boundaries of traditional drone target detection, but also paves the way for a new era of intelligent real-time monitoring in dynamic and complex environments, enabling it to play a meaningful role in practical applications. Specifically, we propose the Hierarchical Extended Path Aggregation Network (HEPAN), which can better fuse features at different levels, thereby improving the model's ability to capture small targets. In addition, we design a C2fDCB lightweight module, which reduces the number of model parameters and computational complexity by optimizing the convolutional structure of the network. The main contributions of this study are as follows:

\begin{itemize}
\item \textbf{Add a head for the small target detection.} Aiming at the problem that YOLOv8 has poor detection effect on small targets in drone images, an additional detection layer for small target detection is added in this paper. This layer significantly enhances the model's ability to capture small targets by fusing shallow and deep feature information.
\item \textbf{Optimize the network structure.} We propose a Hierarchical Extended Path Aggregation Network (HEPAN) to further enhance the ability to fuse features at different levels. In HEPAN,  we add additional convolutional layers in the middle layers of the network structure and use residual connections to strengthen the gradient flow. It significantly enhances the model's ability to capture small objects and reduces the possibility of the detection missing and false detections.
\item \textbf{Introduce the lightweight design.} In this paper, we designs the C2fDCB lightweight module by synthesizing the deep separable convolution~\cite{chollet2017xception} and RepVGG reparameterization method~\cite{ding2021repvgg} to improve the computational efficiency of the C2f module. Meanwhile, the SCDown downsampling module is introduced to reduce the number of model parameters and the computational overhead. Therefore, this model is able to perform efficiently in resource-constrained environments while maintaining high detection accuracy.
\end{itemize}

\begin{figure*}
  \centering
  \begin{subfigure}{0.49\linewidth}
    \includegraphics[width=\textwidth]{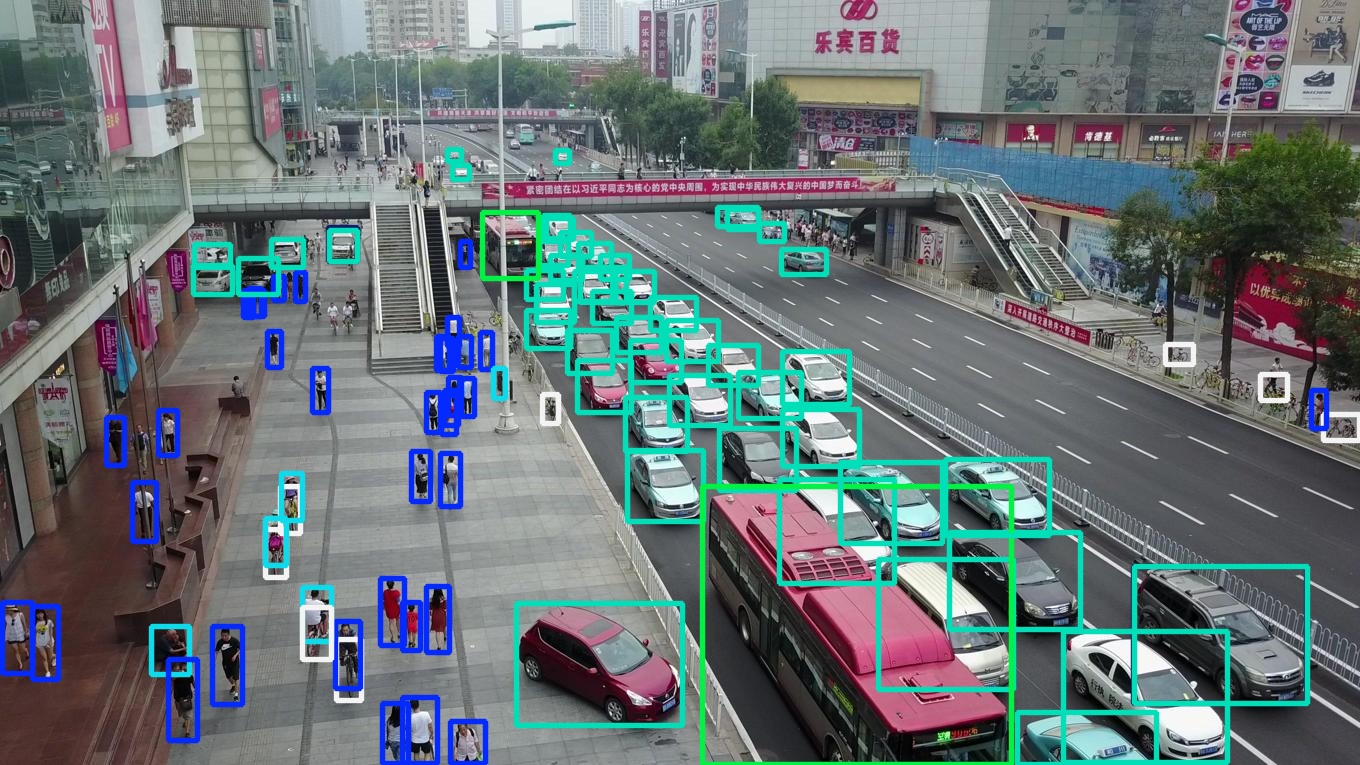}
    \caption{YOLOv8s}
    \label{fig:short-a}
  \end{subfigure}
  \hfill
  \begin{subfigure}{0.49\linewidth}
    \includegraphics[width=\textwidth]{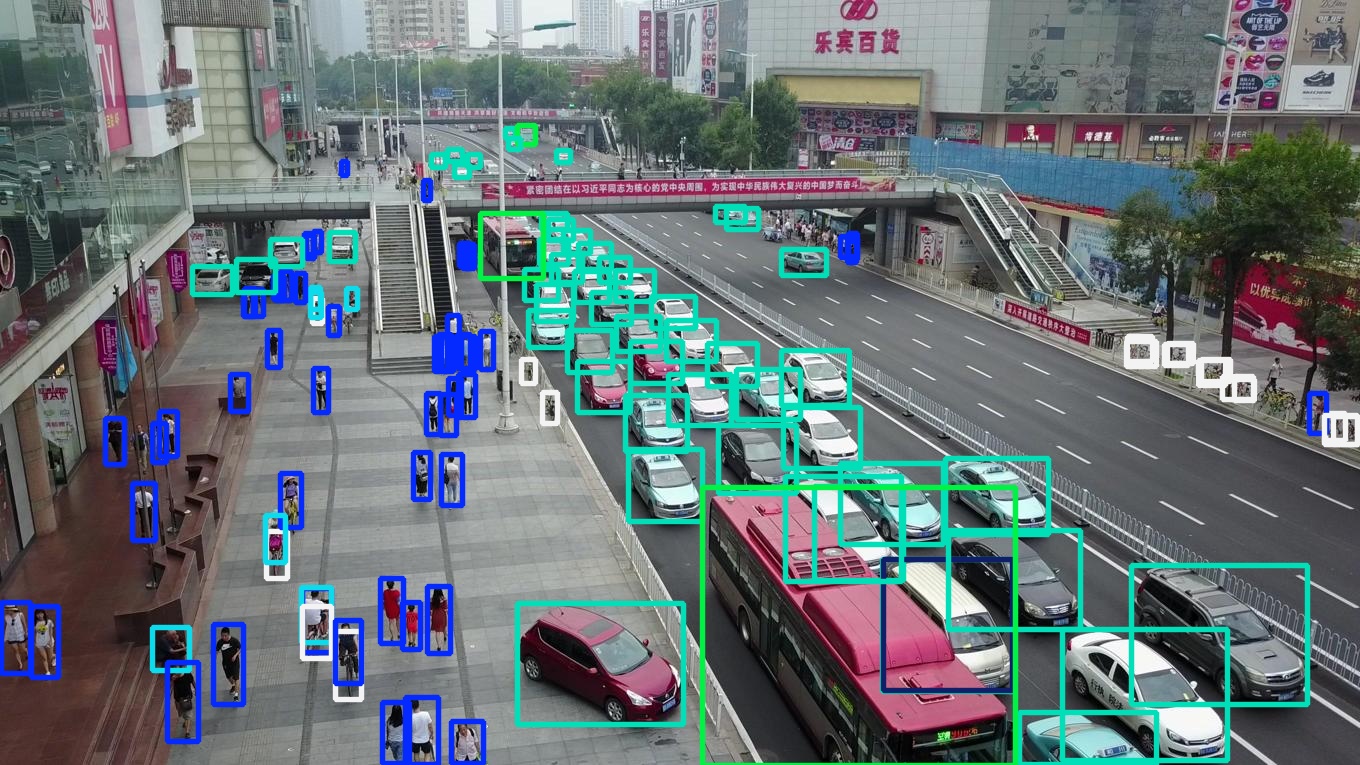}
    \caption{SL-YOLO}
    \label{fig:short-b}
  \end{subfigure}
  \caption{Comparison of detection results between YOLOv8s and SL-YOLO model.}
  \label{fig:short}
\end{figure*}

Through experimental verification on the VisDrone2019 dataset~\cite{du2019visdrone}, our SL-YOLO model proposed in this paper has significantly improved the mAP and other key evaluation indicators, proving that the model is able to achieve good detection performance under limited resource conditions. In Figure~\ref{fig:short}, we demonstrate the performance comparison of the small target detection between YOLOv8s (Fig.~\ref{fig:short}(a)) and our SL-YOLO (Fig.~\ref{fig:short}(b)) is demonstrated. It can be clearly seen that SL-YOLO can more accurately identify small targets in complex backgrounds, and significantly reducing the detection missing and false detections, and hence verifying the effectiveness of our SL-YOLO model.

\section{Related Work}
\label{sec:relat}

One of the difficulties in target detection is to detect targets of different scales, especially small targets. 
Small targets have less feature information and are easily lost during the pooling and downsampling process in deep convolutional neural networks. 
To this end, researchers have proposed multi-scale feature fusion technology to ensure that the network can effectively detect targets of different sizes. 
Early object detection models, such as RCNN~\cite{girshick2014rich} and Fast RCNN~\cite{girshickICCV15fastrcnn}, usually use single-scale feature maps, resulting in limited detection performance. 
Faster RCNN~\cite{renNIPS15fasterrcnn} introduces a Region Proposal Network (RPN), but still relies on fixed-scale features. 
Feature Pyramid Network (FPN)~\cite{lin2017feature} achieves multi-scale detection through multi-resolution feature maps, enhancing small target detection and becoming a classic in multi-scale feature fusion. Path Aggregation Network (PANet)~\cite{liu2018path} improves FPN by adding a bottom-up path for stronger fusion of lower and higher-level features. Adaptively spatial feature fusion (ASFF)~\cite{liu2019learning} refines feature selection by adaptively choosing representative spatial features at each scale. NAS-FPN~\cite{ghiasi2019fpn} uses neural architecture search to automatically identify the optimal fusion strategy, though this method significantly increases computational costs. The weighted bi-directional feature pyramid network (BiFPN)~\cite{tan2020efficientdet} improves on PANet by incorporating a learnable weighting mechanism to balance the contributions of different feature levels, achieving a more effective fusion process.

In recent years, there are some important progresses in the lightweight network design, providing effective solutions for deep learning applications in resource-constrained environments. 
ResNet~\cite{he2016deep} solves the vanishing gradient problem through residual connections; 
DenseNet~\cite{huang2017densely} promotes feature reuse through dense connections, thereby reducing model parameters and improving performance; 
ResNeXt~\cite{xie2017aggregated} introduces the "grouped volumes" and "product" concepts, emphasizes structural diversity and improves model performance by increasing the cardinality. 
MobileNet~\cite{howard2017mobilenets} uses depth-separable convolution to significantly reduce the amount of calculation and is suitable for mobile devices and other scenarios; 
ShuffleNet~\cite{ma2018shufflenet} combines channel shuffling and group convolution to achieve efficient feature learning; 
EfficientNet~\cite {tan2019efficientnet} optimizes the width, depth and resolution of the network through compound scaling, becoming the benchmark for lightweight design. 
CSPNet~\cite{wang2020cspnet} reduces the amount of calculation and improves model expression capabilities through feature map separation and cross-stage connection; 
GhostNet~\cite{han2020ghostnet}'s "Ghost" module further improves feature representation capabilities; 
RepVGG~\cite{ding2021repvgg} significantly reduces the computational overhead of inference through module conversion in the inference stage. 
The design concepts of these networks learn from each other and innovate, promoting the development of practical applications of deep learning models. 

YOLO (You Only Look Once), as a real-time target detection model, has become one of the important benchmarks of computer vision since it was proposed in 2016. 
YOLOv1~\cite{redmon2016you,long2015fully} treats target detection as a regression problem and directly predicts bounding boxes and category labels through a single neural network, significantly improving detection speed, but its small target detection capabilities are limited. 
YOLOv2~\cite{redmon2017yolo9000} introduces multi-scale detection and anchor frame mechanisms to improve model accuracy; 
YOLOv3~\cite{farhadi2018yolov3} uses a residual network structure to enhance small target detection capabilities and achieve a balance between speed and accuracy. 
YOLOv4~\cite{bochkovskiy2020yolov4,zheng2020distance,ghiasi2018dropblock,yun2019cutmix} combines CSPDarknet and a variety of data enhancement technologies to further improve detection performance. 
The later improved versions (YOLOv5, YOLOv6~\cite{li2022yolov6}, YOLOv7~\cite{wang2023yolov7}, YOLOv8 ~\cite{Jocher_Ultralytics_YOLO_2023}, YOLOv9~\cite{wang2024yolov9}, and YOLOv10~\cite{wang2024yolov10}) innovated in ease of use and model lightweightness, making the model widely used in many fields such as autonomous driving, security monitoring, and medical image analysis. 
Future research will continue to focus on improving the robustness and accuracy of the model to cope with more complex scenarios and task requirements. 

\section{Methodology}

\begin{figure*}[h]
\includegraphics[width=1\linewidth]{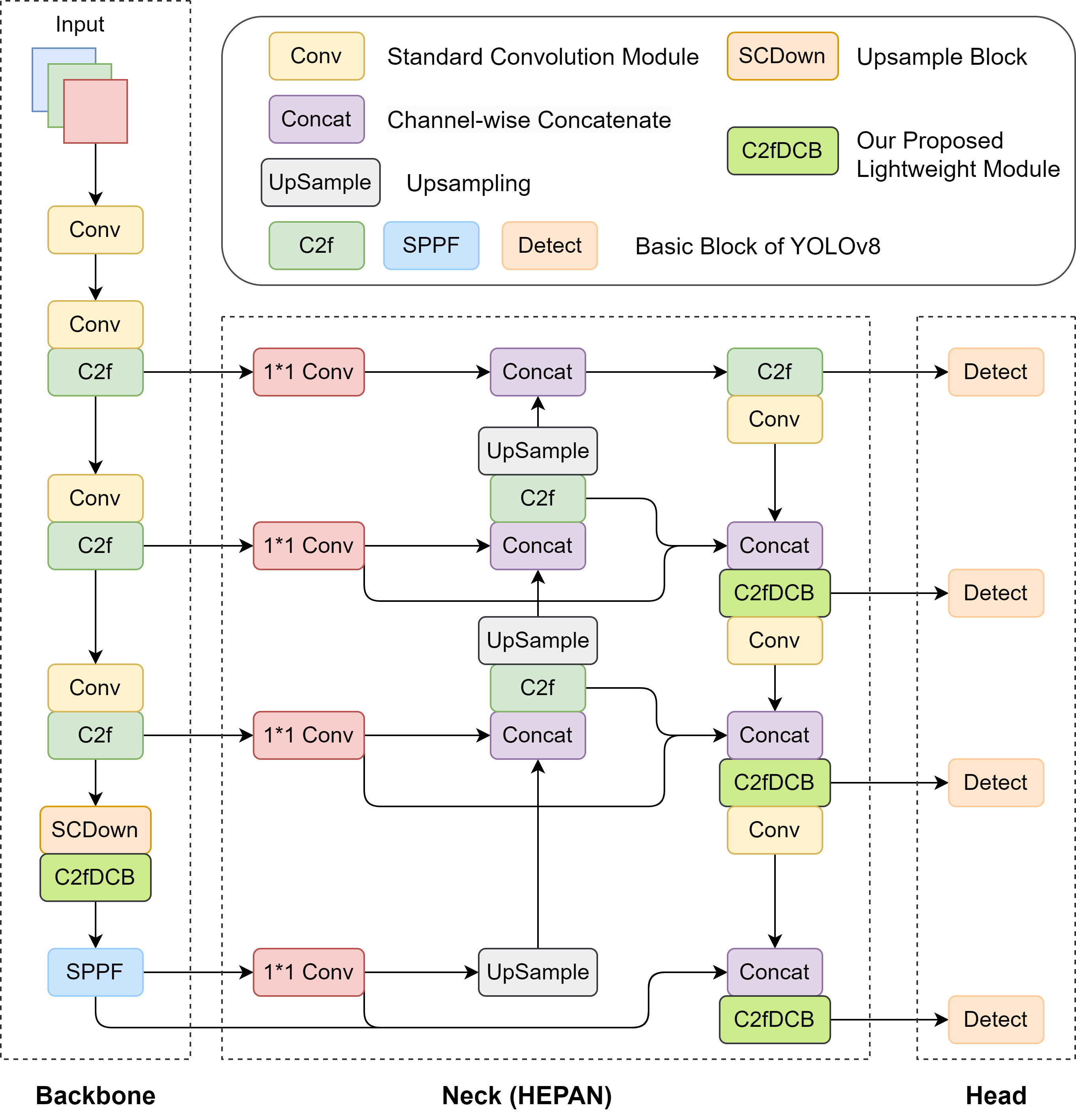}
\caption{The overall structure of our SL-YOLO model.\label{fig4}}
\end{figure*}

In this section, we present the details of the network design of our SL-YOLO model. 
Our goal is to improve the detection performance of the YOLOv8s model, especially on the challenging task of small object detection in drone imagery. 
We push the limits of small object detection by focusing on optimizing the multi-scale feature fusion mechanism and introducing lightweight modules. 
These innovations enable the model to better capture the key feature information of small objects while maintaining excellent accuracy in complex and cluttered backgrounds. 
In the following subsections, we will dive into the core aspects of these enhancements and explore their detailed implementation. 
The overall structure of our enhanced network is shown in Fig.~\ref{fig4}. It consists of three main components: the Backbone, the Neck, and the Head. The Backbone includes standard convolution (Conv), C2f, and our lightweight C2fDCB modules, responsible for feature extraction and compression. The Neck uses the Hierarchical Extended Path Aggregation Network (HEPAN) to fuse multi-scale features, enhancing small object detection. The Head contains four detection layers of different scales to ensure that the model can accurately identify objects at multiple scales.

\subsection{Add a head for the small target detection}

In this study, we firstly applied a common improvement method to solve the problem of YOLOv8's poor detection effect on small targets in drone aerial images - adding a small target detection layer. 
In UAV aerial images with many small targets, although YOLOv8 performs well in common object detection scenarios, its convolutional feature extraction mechanism faces significant challenges when dealing with small targets. 
As the depth of the network increases, the information of small objects is gradually lost in deep features. 
To this end, we introduce a new small object detection layer to the YOLOv8 model, aiming to enhance the detection capabilities of small objects by integrating shallow and deep features. 
Specifically, we upsample in the neck structure of the network to generate a higher resolution (160×160) feature map, and fuse it with the output of the backbone network to improve the feature capturing ability of small targets. 
This method is able to improve the detection effect of small targets and enhance the adaptability of the model.

\subsection{Optimized network structure}

\begin{figure}[h]
\includegraphics[scale=0.46]{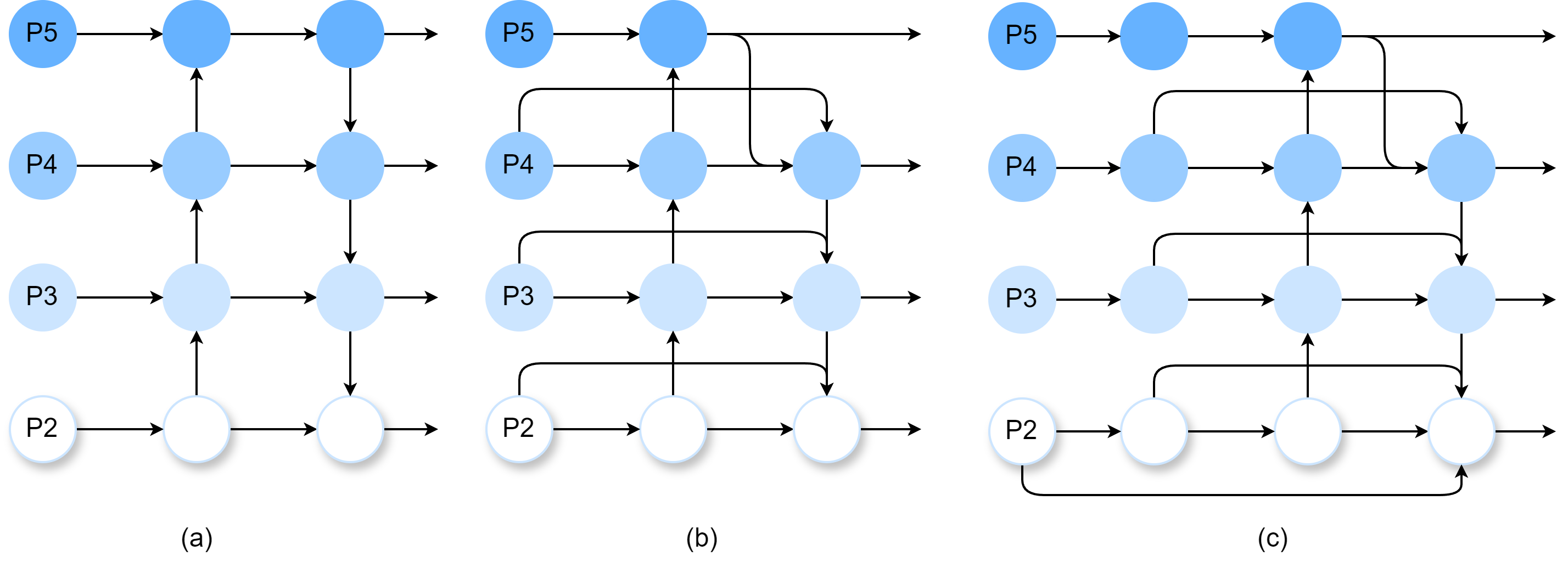}
\caption{The schematic diagrams of network structures: (a) PANet; (b) BiFPN; (c) HEPAN.\label{fig2}}
\end{figure}

The overall structure of YOLOv8 adopts the Path Aggregation Network (PANet), as shown in Fig.~\ref{fig2}(a). 
The backbone network of this structure contains multiple convolutional layers, which gradually increases the depth and resolution of the feature map, allowing the model to effectively capture different levels of feature information. 
However, in practical applications, YOLOv8 shows certain deficiencies when dealing with small objects. 
The main reason is that the effect of feature fusion is not ideal, and the integration of low-level features (such as detailed information of small objects) and high-level features (such as global context information) is insufficient, resulting in limited accuracy and recall of the model in small object detection. 
This limitation significantly affects the model's performance in complex scenes, especially when the scenes have large amount, but relatively small proportion of the small objects.

To address these limitations, Li~\cite{tan2020efficientdet} proposed the weighted bi-directional feature pyramid network (BiFPN), as shown in Fig.~\ref{fig2}(b). 
The core of the BiFPN structure lies in its design of two-way information flow, allowing the model to simultaneously utilize features from different scales. 
Specifically, this structure enhances the interaction between low-level features and high-level features through cross-layer connection and fusion operations, and improves the feature expression ability of small objects. 
In addition, BiFPN ensures that more important features dominate during fusion by adaptively weighting features at different levels during the feature fusion process. 
This design not only improves the detection capabilities of small objects, but also improves the overall performance of the model in complex scenes.

We further optimize the feature fusion mechanism and achieve more detailed feature connection and information flow by introducing the efficient Hierarchical Extended Path Aggregation Network (HEPAN), as shown in Fig.~\ref{fig2}(c). 
In the structure of HEPAN, additional convolutional layers are imported to the neck to further enhance feature extraction and expression capabilities, and residual connections are also be introduced to improve the stability of the gradient flow.
Compared with the traditional PAN and BiFPN structures, HEPAN can significantly improve the accuracy of the small target detection, especially in complex background environments. 

\subsection{Improved Lightweight module}

In the small object detection tasks, the computational complexity and parameter size of the model have a crucial impact on real-time performance and efficiency, especially when deployed on resource-constrained devices. 
Therefore, the lightweight design of the model has become a key link in improving detection speed and reducing power consumption. 
We optimize the design of the model in terms of lightweightness, aiming to reduce redundant calculations while maintaining the excellent detection performance. 
By designing a new feature extraction module, the number of model parameters and computational cost are significantly reduced.

\subsubsection{C2fDCB}

The C2f module in YOLOv8 establishes the connections between feature maps through channel feature fusion technology, effectively integrating feature information from different layers. 
However, there is still room for improvement in parameter efficiency and reducing the computational complexity. 
We design the C2fDCB module by combining the depthwise separable convolution with the classic C2f module to further reduce the parameters and computational complexity while maintaining the richness of features. 
Compared with the traditional convolution layer, this design enables the model to be lightweight in small target detection tasks while still being able to capture key features, making it suitable for resource-constrained drone environments. 
Specifically, without considering bias, the number of parameters and computational complexity of traditional convolution operations can be calculated by formulas (1) and (2), respectively. 
The number of parameters and complexity of depthwise separable convolution are lower than those of traditional convolution, which can be calculated by formulas as (3) and (4), respectively.
\begin{gather}
P_{conv}=K_{1}^2\cdot C_{in}\cdot C_{out} \\
F_{conv}=K_{1}^2\cdot C_{in}\cdot C_{out}\cdot H\cdot W \\
P_{dconv}=K_{2}^2\cdot C_{in}+K_{3}^2\cdot C_{in}\cdot C_{out} \\
F_{dconv}=K_{2}^2\cdot C_{in}\cdot H\cdot W+K_{3}^2\cdot C_{in}\cdot C_{out}\cdot H\cdot W
\end{gather}
where $K_1=3$ is the kernel size for the standard convolution, and $K_2=3$ and $K_3=1$ are the kernel sizes for the depthwise convolution and pointwise convolution, respectively. $C_{\rm in}$ and $C_{\rm out}$ are the numbers of input and output channels, respectively, and $H$ and $W$ are the height and width of the output feature map. 

\begin{figure*}[h]
\centering
\includegraphics[width=1\linewidth]{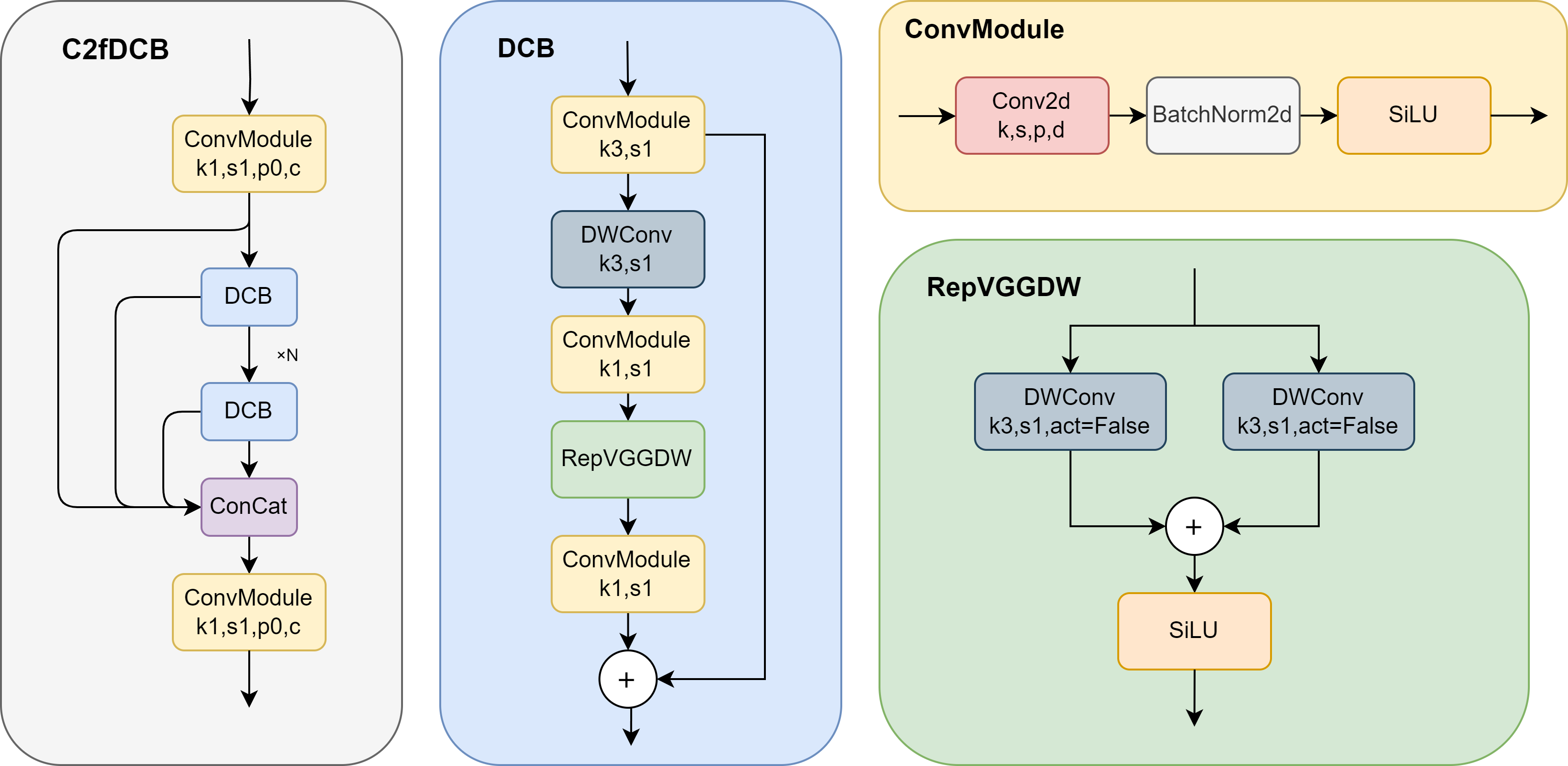}
\caption{The overall structure of the C2fDCB module.\label{fig3}}
\end{figure*}

Depthwise Convolution Block (DCB) is the core component of the C2fDCB module, as shown in Fig.~\ref{fig3}. 
It shows the structure of the C2fDCB module in SL-YOLO, including the key components such as depthwise separable convolution and RepVGGDW convolution. 
The DCB module starts with a 3×3 standard convolution, which is used to maintain the model's feature extraction ability. 
Standard convolution enables the model to capture local patterns and structural information in the input features by applying multiple convolution kernels, laying the foundation for subsequent feature processing. 
Next, the module introduces channel-by-channel convolution, which applies convolution kernels independently to each input channel, effectively reducing the computational complexity. 
Subsequently, point convolution is used to integrate features from different channels. 
Through 1×1 convolution operations, point convolution can effectively mix information from each channel, thereby improving the richness and expressiveness of feature representation. 
The design of this layer ensures the flow of information between channels, allowing the model to learn more complex features. 
Finally, the RepVGGDW convolution is introduced in the module. 
This convolution combines the advantages of depthwise separable convolution, further reducing the computational complexity while maintaining high-performance feature extraction capabilities.

\subsubsection{SCDown}
The SCDown module consists of two main convolutional layers, which aims to efficiently reduce the spatial and channel dimensions of feature maps. 
The first convolution layer uses a 1×1 convolution kernel to compress the number of channels of the input feature map from c1 to c2. 
This process not only reduces the complexity of subsequent calculations, but also allows the model to focus on more critical feature information. 
The second convolution layer further processes the channel-compressed feature map. 
This layer uses a k×k convolution kernel and stride s, and implements channel-wise convolution to improve computational efficiency. 
By adjusting the spatial dimension, the feature map can be effectively downsampled, enhancing the model's ability to capture information at different scales.

\section{Experiments}

\subsection{Dataset}

In this paper, we use the VisDrone2019 dataset~\cite{du2019visdrone}, a large-scale drone-view dataset developed by Tianjin University and other teams. It is primarily used for target detection and contains images with corresponding annotations, divided into training (6471 images), validation (548 images), test (1610 images), and competition sets (1580 images). The image sizes range from 2000 × 1500 to 480 × 360. Due to the drone perspective, the dataset differs from ground-level datasets like MS COCO~\cite{lin2014microsoft} and VOC2012~\cite{everingham2015pascal} in terms of angle, content, background, and lighting. It covers scenes such as streets, parks, and schools under various lighting conditions, and annotates 10 target types, including pedestrians, cars, and bicycles. 

\subsection{Experimental environment}

In this experiment, we chose Ubuntu 24.04.1 as the operating system, and the computing environment used Python 3.10.14, PyTorch 2.3.1, and Cuda 11.8. 
In terms of hardware, we used the NVIDIA RTX 6000 Ada graphics card. 
The implementation of the neural network is based on the YOLOv8 official code provided by Ultralytics, and has been modified accordingly. 
To ensure the repeatability of the experiment, the hyperparameters used in the training, testing, and verification processes are consistent. 
The specific setting is to train 600 epochs, and the size of the input image is adjusted to 640×640. 
All networks are trained using the pre-trained weights yolov8s.pt, and a single image is used for speed measurement (bs=1) during testing.

\subsection{Comparison with the other Models}

Table~\ref{table1} shows the experimental results of various YOLO versions on the VisDrone2019-val dataset, including mAP$_{0.5}(\%)$, mAP$_{0.5:0.95}(\%)$, number of parameters (M), amount of computation (GFLOPs), and single-image inference speed (FPS). 
From the results in the table, it can be seen that SL-YOLO performs well in multiple indicators, especially in mAP and inference speed. 
Specifically, SL-YOLO achieved 46.9\% and 28.9\% in mAP$_{0.5}$ and mAP$_{0.5:0.95}$, respectively, with higher detection accuracy than other models. 
Compared with the standard YOLOv8s (43.0\%, 26.0\%), SL-YOLO's mAP increased by nearly 4 percentage points, showing its significant advantage in accuracy. 
In addition, SL-YOLO also performs well in inference speed, with an FPS of 132, close to lighter models such as YOLOv8s (163) and YOLOv8s-p2 (139), while its performance advantage is reflected in its lower parameter count (9.6M) and computational complexity (36.7 GFLOPs), demonstrating its efficient computing and inference capabilities.

\begin{table*}[htbp!]
\centering
\begin{tabular}{ccccccccc}
\toprule
\textbf{Model} & \textbf{\boldmath mAP$_{0.5}(\%)$} & \textbf{\boldmath mAP$_{0.5:0.95}(\%)$} & \textbf{params(M)} & \textbf{GFLOPs} & \textbf{\boldmath FPS$_{bs=1}$}\\
\midrule
yolov8s & 43.0 & 26.0 & 11.1 & 28.5 & 163 \\
yolov8m & 46.7 & 28.9 & 25.8 & 78.7 & 133 \\
yolov8s-p2 & 46.2 & 28.3 & 10.6 & 36.7 & 139 \\
yolov9s & 42.1 & 25.3 & 7.2 & 26.7 & 77 \\
yolov9m & 47.0 & 29.4 & 20.0 & 76.5 & 92 \\
yolov10s & 41.2 & 24.8 & 8.0 & 24.5 & 110 \\
yolov10s-p2 & 44.8 & 27.9 & 8.2 & 36.6 & 98 \\
yolov10m & 45.3 & 28.1 & 16.5 & 63.5 & 92 \\
yolov11s & 41.6 & 25.2 & 9.4 & 21.3 & 128 \\
yolov11s-p2 & 45.4 & 28.0 & 9.6 & 29.2 & 114 \\
yolov11m & 47.3 & 29.3 & 20.0 & 67.7 & 108 \\
\rowcolor[gray]{0.9}
SL-YOLO & 46.9 & 28.9 & 9.6 & 36.7 & 132 \\
\bottomrule
\end{tabular}
\caption{Experimental results compared with other models in VisDrone2019-val.\label{table1}}
\end{table*}

\begin{table}[htbp!]
\centering
\begin{tabular}{ccccccccc}
\toprule
\textbf{Model} & \textbf{\boldmath mAP$_{0.5}(\%)$} & \textbf{params(M)} & \textbf{GFLOPs} \\
\midrule
yolov8s & 35.2 & 11.1 & 28.5 \\
yolov8s-p2 & 37.4 & 10.6 & 36.7 \\
yolov8m & 38.2 & 25.8 & 78.7 \\
yolov9s & 34.3 & 7.2 & 26.7 \\
yolov9m & 38.9 & 20.0 & 76.5 \\
yolov10s & 33.7 & 8.0 & 24.5 \\
yolov10s-p2 & 36.4 & 8.0 & 36.3 \\
yolov10m & 37.8 & 16.5 & 63.5 \\
yolov11s & 33.8 & 9.4 & 21.3 \\
yolov11s-p2 & 36.7 & 9.6 & 29.2 \\
yolov11m & 38.8 & 20.0 & 67.7 \\
\rowcolor[gray]{0.9}
SL-YOLO & 38.3 & 9.6 & 36.7 \\
\bottomrule
\end{tabular}
\caption{Experimental results compared with other models in VisDrone2019-test.\label{table2}}
\end{table}

Table~\ref{table2} shows the experimental results of various YOLO versions on the VisDrone2019-test dataset. 
From the data in the table, it can be seen that SL-YOLO's mAP$_{0.5}$ reached 38.3\%, which is in a leading position compared with other YOLO models, only slightly lower than YOLOv9m (38.9\%) and YOLOv11m (38.8\%), but its parameter volume (9.6M) and computational complexity (36.7 GFLOPs) are much lower than these two (20.0M and 67.7 GFLOPs, respectively). 
In contrast, SL-YOLO's computational efficiency is more outstanding, with the same computational complexity as YOLOv8s-p2, but higher accuracy (37.4\% vs 38.3\%). 
In addition, SL-YOLO also has an advantage in parameter volume, which is almost the same as YOLOv8s-p2 (10.6M), but its accuracy has been significantly improved.

In summary, SL-YOLO has found a good balance between accuracy and speed. 
It is one of the most efficient and excellent models in this task, and is especially suitable for scenarios with high requirements for real-time performance and accuracy.

\begin{table*}[htbp!]
\begin{tabular}{ccccccccc}
\toprule
\textbf{Baseline} & \textbf{p2} & \textbf{HEPAN} & \textbf{C2fDCB} & \textbf{SCDown} & \textbf{\boldmath mAP$_{0.5}(\%)$} & \textbf{\boldmath mAP$_{0.5:0.95}(\%)$} & \textbf{params(M)} & \textbf{GFLOPs} \\
\midrule
\ding{51} &  &  &  &  & 43.0 & 26.0 & 11.1 & 28.5 \\
\ding{51} & \ding{51} &  &  &  & 46.2 & 28.3 & 10.6 & 36.7 \\
\ding{51} & \ding{51} & \ding{51} &  &  & 47.2 & 29.1 & 11.3 & 38.1 \\
\ding{51} & \ding{51} & \ding{51} & \ding{51} &  & 47.0 & 29.1 & 10.6 & 37.3 \\
\rowcolor[gray]{0.9}
\ding{51} & \ding{51} & \ding{51} & \ding{51} & \ding{51} & 46.9 & 28.9 & 9.6 & 36.7 \\
\bottomrule
\end{tabular}
\caption{Ablation experiment result in VisDrone2019-val.\label{table3}}
\end{table*}

\subsection{Ablation Study}
In order to further verify the effectiveness of the proposed algorithm, an ablation experiment was performed on the VisDrone-2019 validation set. 
Taking YOLOv8s as the baseline model, the various improvement methods mentioned in this article are gradually added in combination to verify the improvement of target detection performance of each method. 
Table~\ref{table3} gives the results of ablation experiments performed on the VisDrone-2019 validation dataset. 
The experiment involves adding various improvements to the basic YOLOv8s model, including adding a P2 layer small target detection head, HEPAN structure, C2fDCB module, SCDown and combinations of these improvements. 
It can be seen from the table that the basic YOLOv8s model obtained mAP$_{0.5}$ in the experiment of 43.0\%, mAP$_{0.5:0.95}$ of 26.0\%, the number of parameters was 11.1M, and the GFLOPs were 28.5. 
After adding the P2 layer to the Baseline, mAP$_{0.5}$ increased to 46.2\% (+3.2\%), mAP$_{0.5:0.95}$ increased to 28.3\% (+2.3\%), and the number of parameters was reduced from 11.1M to 10.6M, GFLOPs increased to 36.7. 
It shows that adding a small target layer significantly improves the detection accuracy, especially the detection ability of small targets, but the amount of calculation increases. 
After further improvement to the HEPAN structure, mAP$_{0.5}$ further increased to 47.2\% (+1.0\%), mAP$_{0.5:0.95}$ increased to 29.1\% (+0.8\%), and the number of parameters increased to 11.3M, GFLOPs is 38.1. 
This result demonstrates that the HEPAN structure has further improved the overall network performance, especially the performance of high IoU thresholds. 
After replacing the C2f module with C2fDCB, mAP$_{0.5}$ dropped slightly to 47.0\% (-0.2\%), mAP$_{0.5:0.95}$ remained at 29.1\%, the number of parameters was reduced to 10.6M, and GFLOPs were also reduced to 37.3. 
Although the accuracy decreased slightly, the model became lighter, and the amount of parameters and calculations was reduced. 
Finally, after the downsampling module was replaced with the SCDown module, the mAP$_{0.5}$ of the model was 46.9\%, mAP$_{0.5:0.95}$ was 28.9\%, and the number of parameters was significantly reduced to 9.6 M (-1.0 M), GFLOPs reduced to 36.7 (-0.6), showing the advantages of the SCDown module in simplifying the model structure.

\textbf{Analysis of different network structures.} In the Table~\ref{table4}, the model performances by using the three network structures of PAN, BiFPN and HEPAN on the VisDrone2019-val dataset are compared. 
The HEPAN structure achieved the highest detection effect, reaching mAP$_{0.5}$(47.2\%), with a parameter size of 11.3M and GFLOPs of 38.1, which is better than the PAN and BiFPN structures (46.2\% and 46.4\%, respectively). 
Although the computational complexity of HEPAN is slightly higher, the performance improvement is significant.

\begin{table}[htbp!]
\centering
\begin{tabular}{ccccccccc}
\toprule
\textbf{Structure} & \textbf{\boldmath mAP$_{0.5}(\%)$} & \textbf{params(M)} & \textbf{GFLOPs} \\
\midrule
PAN & 46.2 & 10.6 & 36.7 \\
BiFPN & 46.4 & 10.7 & 37.1 \\
\rowcolor[gray]{0.9}
HEPAN & 47.2 & 11.3 & 38.1 \\
\bottomrule
\end{tabular}
\caption{Experiment result of different network structures in VisDrone2019-val.\label{table4}}
\end{table}

\textbf{Analysis of different modules.} Table~\ref{table5} compares the effects of using the modules of C3, C2f, and C2fDCB, respectively. 
The C2f module achieves the highest mAP$_{0.5}$ (47.2\%), with 11.3M parameters and 38.1 GFLOPs. 
In contrast, C2fDCB has slightly lower parameters and computation (10.6M and 37.3 GFLOPs), but its performance is slightly lower, reaching mAP$_{0.5}$ of 47.0\%. 
This shows that the C2f module has a good balance between the performance and computation.

\begin{table}[htbp!]
\centering
\begin{tabular}{ccccccccc}
\toprule
\textbf{Module} & \textbf{\boldmath mAP$_{0.5}(\%)$} & \textbf{params(M)} & \textbf{GFLOPs} \\
\midrule
C3 & 46.2 & 9.8 & 36.0 \\
C2f & 47.2 & 11.3 & 38.1 \\
\rowcolor[gray]{0.9}
C2fDCB & 47.0 & 10.6 & 37.3 \\
\bottomrule
\end{tabular}
\caption{Experiment result of different modules in VisDrone2019-val.\label{table5}}
\end{table}

\textbf{Analysis of different downsampling modules.} The experimental results are shown in Table \ref{table6}. Although the Conv module is slightly better than SCDown in mAP$_{0.5}$, the performance gap between the two is almost negligible. 
However, the SCDown module is better than Conv in terms of both parameter quantity and computational complexity. 
Therefore, choosing the SCDown module may bring a better balance in practical applications, especially when resources are limited, and still maintain good detection performance.

\begin{table}[htbp!]
\centering
\begin{tabular}{ccccccccc}
\toprule
\textbf{Module} & \textbf{\boldmath mAP$_{0.5}(\%)$} & \textbf{params(M)} & \textbf{GFLOPs} \\
\midrule
Conv & 47.0 & 10.6 & 37.3 \\
\rowcolor[gray]{0.9}
SCDown & 46.9 & 9.6 & 36.7 \\
\bottomrule
\end{tabular}
\caption{Experiment result of different downsampling modules in VisDrone2019-val.\label{table6}}
\end{table} 

\section{Conclusions}
In this paper, we present the SL-YOLO model, a stronger and lighter model for drone target detection in complex environments. SL-YOLO incorporates a Hierarchical Extended Path Aggregation Network (HEPAN) for improved cross-scale feature fusion, enhancing small target accuracy, along with the C2fDCB and SCDown modules to reduce parameters and computational load while maintaining high performance. The design of SL-YOLO enables drones to efficiently detect small targets in complex environments, providing solid support for key tasks such as disaster response and intelligent monitoring. 
In the future, we plan to further optimize SL-YOLO's cross-scenario adaptability so that it can demonstrate excellent small target detection performance in more different scenarios, providing more possibilities for the intelligent development of the drone field. 

{
    \small
    \bibliographystyle{ieeenat_fullname}
    \bibliography{main}

\begin{thebibliography}{41}
\providecommand{\natexlab}[1]{#1}
\providecommand{\url}[1]{\texttt{#1}}
\expandafter\ifx\csname urlstyle\endcsname\relax
  \providecommand{\doi}[1]{doi: #1}\else
  \providecommand{\doi}{doi: \begingroup \urlstyle{rm}\Url}\fi

\bibitem[Bochkovskiy et~al.(2020)Bochkovskiy, Wang, and Liao]{bochkovskiy2020yolov4}
Alexey Bochkovskiy, Chien-Yao Wang, and Hong-Yuan~Mark Liao.
\newblock Yolov4: Optimal speed and accuracy of object detection, 2020.

\bibitem[Chollet(2017)]{chollet2017xception}
Fran{\c{c}}ois Chollet.
\newblock Xception: Deep learning with depthwise separable convolutions.
\newblock In \emph{CVPR}, pages 1251--1258, 2017.

\bibitem[Ding et~al.(2021)Ding, Zhang, Ma, Han, Ding, and Sun]{ding2021repvgg}
Xiaohan Ding, Xiangyu Zhang, Ningning Ma, Jungong Han, Guiguang Ding, and Jian Sun.
\newblock Repvgg: Making vgg-style convnets great again.
\newblock In \emph{CVPR}, pages 13733--13742, 2021.

\bibitem[Du et~al.(2019)Du, Zhu, Wen, Bian, Lin, Hu, Peng, Zheng, Wang, Zhang, et~al.]{du2019visdrone}
Dawei Du, Pengfei Zhu, Longyin Wen, Xiao Bian, Haibin Lin, Qinghua Hu, Tao Peng, Jiayu Zheng, Xinyao Wang, Yue Zhang, et~al.
\newblock Visdrone-det2019: The vision meets drone object detection in image challenge results.
\newblock In \emph{ICCV}, pages 0--0, 2019.

\bibitem[Everingham et~al.(2015)Everingham, Eslami, Van~Gool, Williams, Winn, and Zisserman]{everingham2015pascal}
Mark Everingham, SM~Ali Eslami, Luc Van~Gool, Christopher~KI Williams, John Winn, and Andrew Zisserman.
\newblock The pascal visual object classes challenge: A retrospective.
\newblock \emph{IJCV}, 111:\penalty0 98--136, 2015.

\bibitem[Farhadi and Redmon(2018)]{farhadi2018yolov3}
Ali Farhadi and Joseph Redmon.
\newblock Yolov3: An incremental improvement.
\newblock In \emph{CVPR}, pages 1--6. Springer Berlin/Heidelberg, Germany, 2018.

\bibitem[Ghiasi et~al.(2018)Ghiasi, Lin, and Le]{ghiasi2018dropblock}
Golnaz Ghiasi, Tsung-Yi Lin, and Quoc~V Le.
\newblock Dropblock: A regularization method for convolutional networks.
\newblock \emph{NeurIPS}, 31, 2018.

\bibitem[Ghiasi et~al.(2019)Ghiasi, Lin, and Le]{ghiasi2019fpn}
Golnaz Ghiasi, Tsung-Yi Lin, and Quoc~V Le.
\newblock Nas-fpn: Learning scalable feature pyramid architecture for object detection.
\newblock In \emph{CVPR}, pages 7036--7045, 2019.

\bibitem[Girshick(2015)]{girshickICCV15fastrcnn}
Ross Girshick.
\newblock Fast r-cnn.
\newblock In \emph{ICCV}, 2015.

\bibitem[Girshick et~al.(2014)Girshick, Donahue, Darrell, and Malik]{girshick2014rich}
Ross Girshick, Jeff Donahue, Trevor Darrell, and Jitendra Malik.
\newblock Rich feature hierarchies for accurate object detection and semantic segmentation.
\newblock In \emph{CVPR}, pages 580--587, 2014.

\bibitem[Han et~al.(2020)Han, Wang, Tian, Guo, Xu, and Xu]{han2020ghostnet}
Kai Han, Yunhe Wang, Qi Tian, Jianyuan Guo, Chunjing Xu, and Chang Xu.
\newblock Ghostnet: More features from cheap operations.
\newblock In \emph{CVPR}, pages 1580--1589, 2020.

\bibitem[He et~al.(2016)He, Zhang, Ren, and Sun]{he2016deep}
Kaiming He, Xiangyu Zhang, Shaoqing Ren, and Jian Sun.
\newblock Deep residual learning for image recognition.
\newblock In \emph{CVPR}, pages 770--778, 2016.

\bibitem[Howard et~al.(2019)Howard, Sandler, Chu, Chen, Chen, Tan, Wang, Zhu, Pang, Vasudevan, et~al.]{howard2019searching}
Andrew Howard, Mark Sandler, Grace Chu, Liang-Chieh Chen, Bo Chen, Mingxing Tan, Weijun Wang, Yukun Zhu, Ruoming Pang, Vijay Vasudevan, et~al.
\newblock Searching for mobilenetv3.
\newblock In \emph{ICCV}, pages 1314--1324, 2019.

\bibitem[Howard(2017)]{howard2017mobilenets}
Andrew~G Howard.
\newblock Mobilenets: Efficient convolutional neural networks for mobile vision applications.
\newblock \emph{arXiv preprint arXiv:1704.04861}, 2017.

\bibitem[Huang et~al.(2017)Huang, Liu, Van Der~Maaten, and Weinberger]{huang2017densely}
Gao Huang, Zhuang Liu, Laurens Van Der~Maaten, and Kilian~Q Weinberger.
\newblock Densely connected convolutional networks.
\newblock In \emph{CVPR}, pages 4700--4708, 2017.

\bibitem[Jocher et~al.(2023)Jocher, Qiu, and Chaurasia]{Jocher_Ultralytics_YOLO_2023}
Glenn Jocher, Jing Qiu, and Ayush Chaurasia.
\newblock Ultralytics yolo.
\newblock \emph{https://github.com/ultralytics/ultralytics}, 2023.

\bibitem[Li et~al.(2022)Li, Li, Jiang, Weng, Geng, Li, Ke, Li, Cheng, Nie, et~al.]{li2022yolov6}
Chuyi Li, Lulu Li, Hongliang Jiang, Kaiheng Weng, Yifei Geng, Liang Li, Zaidan Ke, Qingyuan Li, Meng Cheng, Weiqiang Nie, et~al.
\newblock Yolov6: A single-stage object detection framework for industrial applications.
\newblock \emph{arXiv preprint arXiv:2209.02976}, 2022.

\bibitem[Lin et~al.(2014)Lin, Maire, Belongie, Hays, Perona, Ramanan, Doll{\'a}r, and Zitnick]{lin2014microsoft}
Tsung-Yi Lin, Michael Maire, Serge Belongie, James Hays, Pietro Perona, Deva Ramanan, Piotr Doll{\'a}r, and C~Lawrence Zitnick.
\newblock Microsoft coco: Common objects in context.
\newblock In \emph{ECCV}, pages 740--755. Springer, 2014.

\bibitem[Lin et~al.(2017)Lin, Doll{\'a}r, Girshick, He, Hariharan, and Belongie]{lin2017feature}
Tsung-Yi Lin, Piotr Doll{\'a}r, Ross Girshick, Kaiming He, Bharath Hariharan, and Serge Belongie.
\newblock Feature pyramid networks for object detection.
\newblock In \emph{CVPR}, pages 2117--2125, 2017.

\bibitem[Liu et~al.(2018)Liu, Qi, Qin, Shi, and Jia]{liu2018path}
Shu Liu, Lu Qi, Haifang Qin, Jianping Shi, and Jiaya Jia.
\newblock Path aggregation network for instance segmentation.
\newblock In \emph{CVPR}, pages 8759--8768, 2018.

\bibitem[Liu et~al.(2019)Liu, Huang, and Wang]{liu2019learning}
Songtao Liu, Di Huang, and Yunhong Wang.
\newblock Learning spatial fusion for single-shot object detection.
\newblock \emph{arXiv preprint arXiv:1911.09516}, 2019.

\bibitem[Long et~al.(2015)Long, Shelhamer, and Darrell]{long2015fully}
Jonathan Long, Evan Shelhamer, and Trevor Darrell.
\newblock Fully convolutional networks for semantic segmentation.
\newblock In \emph{CVPR}, pages 3431--3440, 2015.

\bibitem[Ma et~al.(2018)Ma, Zhang, Zheng, and Sun]{ma2018shufflenet}
Ningning Ma, Xiangyu Zhang, Hai-Tao Zheng, and Jian Sun.
\newblock Shufflenet v2: Practical guidelines for efficient cnn architecture design.
\newblock In \emph{ECCV}, pages 116--131, 2018.

\bibitem[Redmon(2016)]{redmon2016you}
J Redmon.
\newblock You only look once: Unified, real-time object detection.
\newblock In \emph{CVPR}, 2016.

\bibitem[Redmon and Farhadi(2017)]{redmon2017yolo9000}
Joseph Redmon and Ali Farhadi.
\newblock Yolo9000: better, faster, stronger.
\newblock In \emph{CVPR}, pages 7263--7271, 2017.

\bibitem[Ren et~al.(2015)Ren, He, Girshick, and Sun]{renNIPS15fasterrcnn}
Shaoqing Ren, Kaiming He, Ross Girshick, and Jian Sun.
\newblock Faster {R-CNN}: Towards real-time object detection with region proposal networks.
\newblock In \emph{NeurIPS}, 2015.

\bibitem[Sandler et~al.(2018)Sandler, Howard, Zhu, Zhmoginov, and Chen]{sandler2018mobilenetv2}
Mark Sandler, Andrew Howard, Menglong Zhu, Andrey Zhmoginov, and Liang-Chieh Chen.
\newblock Mobilenetv2: Inverted residuals and linear bottlenecks.
\newblock In \emph{CVPR}, pages 4510--4520, 2018.

\bibitem[Tan and Le(2019)]{tan2019efficientnet}
Mingxing Tan and Quoc Le.
\newblock Efficientnet: Rethinking model scaling for convolutional neural networks.
\newblock In \emph{International conference on machine learning}, pages 6105--6114. PMLR, 2019.

\bibitem[Tan and Le(2021)]{tan2021efficientnetv2}
Mingxing Tan and Quoc Le.
\newblock Efficientnetv2: Smaller models and faster training.
\newblock In \emph{International conference on machine learning}, pages 10096--10106. PMLR, 2021.

\bibitem[Tan et~al.(2020)Tan, Pang, and Le]{tan2020efficientdet}
Mingxing Tan, Ruoming Pang, and Quoc~V Le.
\newblock Efficientdet: Scalable and efficient object detection.
\newblock In \emph{CVPR}, pages 10781--10790, 2020.

\bibitem[Tang et~al.(2023)Tang, Ni, Zhao, Gu, and Cao]{tang2023survey}
Guangyi Tang, Jianjun Ni, Yonghao Zhao, Yang Gu, and Weidong Cao.
\newblock A survey of object detection for uavs based on deep learning.
\newblock \emph{Remote Sensing}, 16\penalty0 (1):\penalty0 149, 2023.

\bibitem[Vaswani(2017)]{vaswani2017attention}
A Vaswani.
\newblock Attention is all you need.
\newblock \emph{NeurIPS}, 2017.

\bibitem[Wang et~al.(2024{\natexlab{a}})Wang, Chen, Liu, Chen, Lin, Han, and Ding]{wang2024yolov10}
Ao Wang, Hui Chen, Lihao Liu, Kai Chen, Zijia Lin, Jungong Han, and Guiguang Ding.
\newblock Yolov10: Real-time end-to-end object detection.
\newblock \emph{arXiv preprint arXiv:2405.14458}, 2024{\natexlab{a}}.

\bibitem[Wang et~al.(2020)Wang, Liao, Wu, Chen, Hsieh, and Yeh]{wang2020cspnet}
Chien-Yao Wang, Hong-Yuan~Mark Liao, Yueh-Hua Wu, Ping-Yang Chen, Jun-Wei Hsieh, and I-Hau Yeh.
\newblock Cspnet: A new backbone that can enhance learning capability of cnn.
\newblock In \emph{CVPR}, pages 390--391, 2020.

\bibitem[Wang et~al.(2023)Wang, Bochkovskiy, and Liao]{wang2023yolov7}
Chien-Yao Wang, Alexey Bochkovskiy, and Hong-Yuan~Mark Liao.
\newblock Yolov7: Trainable bag-of-freebies sets new state-of-the-art for real-time object detectors.
\newblock In \emph{CVPR}, pages 7464--7475, 2023.

\bibitem[Wang et~al.(2024{\natexlab{b}})Wang, Yeh, and Liao]{wang2024yolov9}
Chien-Yao Wang, I-Hau Yeh, and Hong-Yuan~Mark Liao.
\newblock Yolov9: Learning what you want to learn using programmable gradient information, 2024{\natexlab{b}}.

\bibitem[Wu et~al.(2021)Wu, Li, Hong, Tao, and Du]{wu2021deep}
Xin Wu, Wei Li, Danfeng Hong, Ran Tao, and Qian Du.
\newblock Deep learning for unmanned aerial vehicle-based object detection and tracking: A survey.
\newblock \emph{IEEE Geoscience and Remote Sensing Magazine}, 10\penalty0 (1):\penalty0 91--124, 2021.

\bibitem[Xie et~al.(2017)Xie, Girshick, Doll{\'a}r, Tu, and He]{xie2017aggregated}
Saining Xie, Ross Girshick, Piotr Doll{\'a}r, Zhuowen Tu, and Kaiming He.
\newblock Aggregated residual transformations for deep neural networks.
\newblock In \emph{CVPR}, pages 1492--1500, 2017.

\bibitem[Yun et~al.(2019)Yun, Han, Oh, Chun, Choe, and Yoo]{yun2019cutmix}
Sangdoo Yun, Dongyoon Han, Seong~Joon Oh, Sanghyuk Chun, Junsuk Choe, and Youngjoon Yoo.
\newblock Cutmix: Regularization strategy to train strong classifiers with localizable features.
\newblock In \emph{ICCV}, pages 6023--6032, 2019.

\bibitem[Zhang et~al.(2018)Zhang, Zhou, Lin, and Sun]{zhang2018shufflenet}
Xiangyu Zhang, Xinyu Zhou, Mengxiao Lin, and Jian Sun.
\newblock Shufflenet: An extremely efficient convolutional neural network for mobile devices.
\newblock In \emph{CVPR}, pages 6848--6856, 2018.

\bibitem[Zheng et~al.(2020)Zheng, Wang, Liu, Li, Ye, and Ren]{zheng2020distance}
Zhaohui Zheng, Ping Wang, Wei Liu, Jinze Li, Rongguang Ye, and Dongwei Ren.
\newblock Distance-iou loss: Faster and better learning for bounding box regression.
\newblock In \emph{AAAI}, pages 12993--13000, 2020.

\end{thebibliography}
}


\end{document}